\documentclass[nohyperref]{article}

\usepackage{microtype}
\usepackage{graphicx}
\usepackage{subfigure}
\usepackage{booktabs} 

\usepackage{hyperref}

\usepackage[accepted]{icml2022_arxiv}

\usepackage{amsmath}
\usepackage{amssymb}
\usepackage{mathtools}
\usepackage{amsthm}

\usepackage[capitalize,noabbrev]{cleveref}

\usepackage{amsmath,amsfonts,bm}

\def\eqref#1{equation~\ref{#1}}

\def\1{\bm{1}}

\def\vb{{\bm{b}}}

\def\vh{{\bm{h}}}

\def\vn{{\bm{n}}}

\def\vp{{\bm{p}}}
\def\vq{{\bm{q}}}

\def\vv{{\bm{v}}}

\def\vx{{\bm{x}}}
\def\vy{{\bm{y}}}

\DeclareMathAlphabet{\mathsfit}{\encodingdefault}{\sfdefault}{m}{sl}
\SetMathAlphabet{\mathsfit}{bold}{\encodingdefault}{\sfdefault}{bx}{n}

\usepackage{hyperref}
\usepackage{url}
\usepackage{mwe}
\usepackage{siunitx}

\theoremstyle{plain}

\theoremstyle{definition}

\theoremstyle{remark}

\usepackage[textsize=tiny]{todonotes}

\icmltitlerunning{Fast Aquatic Swimmer Optimization with Differentiable Projective Dynamics and Neural Network Hydrodynamic Models}

\begin{document}

\twocolumn[
\icmltitle{Fast Aquatic Swimmer Optimization with Differentiable Projective Dynamics and Neural Network Hydrodynamic Models}



\icmlsetsymbol{equal}{*}

\begin{icmlauthorlist}
\icmlauthor{Elvis Nava}{ethai,eth,eth-ni}
\icmlauthor{John Z. Zhang}{eth}
\icmlauthor{Mike Y. Michelis}{eth}
\icmlauthor{Tao Du}{mit}
\icmlauthor{Pingchuan Ma}{mit} 
\icmlauthor{Benjamin F. Grewe}{eth-ni}
\icmlauthor{Wojciech Matusik}{mit}
\icmlauthor{Robert K. Katzschmann}{eth}
\end{icmlauthorlist}

\icmlaffiliation{ethai}{ETH AI Center, ETH Zurich, Zurich, Switzerland}
\icmlaffiliation{eth}{Soft Robotics Lab, ETH Zurich, Zurich, Switzerland}
\icmlaffiliation{eth-ni}{Institute of Neuroinformatics, ETH Zurich, Zurich, Switzerland}
\icmlaffiliation{mit}{CSAIL, MIT, Cambridge, MA, USA}

\icmlcorrespondingauthor{Elvis Nava}{elvis.nava@ai.ethz.ch}
\icmlcorrespondingauthor{Robert K. Katzschmann}{rkk@ethz.ch}

\icmlkeywords{Machine Learning, Robotics, Soft Robotics, Optimization, Physics, Hydrodynamics, Robotic Fish}

\vskip 0.3in
]



\printAffiliationsAndNotice{}  
\begin{abstract}
Aquatic locomotion is a classic fluid-structure interaction (FSI) problem of interest to biologists and engineers. Solving the fully coupled FSI equations for incompressible Navier-Stokes and finite elasticity is computationally expensive. Optimizing robotic swimmer design within such a system generally involves cumbersome, gradient-free procedures on top of the already costly simulation.
To address this challenge we present a novel, fully differentiable hybrid approach to FSI that combines a 2D direct numerical simulation for the deformable solid structure of the swimmer and a physics-constrained neural network surrogate to capture hydrodynamic effects of the fluid.
For the deformable solid simulation of the swimmer's body, we use state-of-the-art techniques from the field of computer graphics to speed up the finite-element method (FEM). For the fluid simulation, we use a U-Net architecture trained with a physics-based loss function to predict the flow field at each time step. The pressure and velocity field outputs from the neural network are sampled around the boundary of our swimmer using an immersed boundary method (IBM) to compute its swimming motion accurately and efficiently.
We demonstrate the computational efficiency and differentiability of our hybrid simulator on a 2D carangiform swimmer. Due to differentiability, the simulator can be used for computational design of controls for soft bodies immersed in fluids via direct gradient-based optimization.
\end{abstract}

\begin{figure*}[ht]
\vskip 0.2in
\begin{center}
\centerline{
\begin{tabular}{c@{\hskip1mm}c@{\hskip1mm}c}
\includegraphics[trim=50 20 570 20,clip,width=56mm]{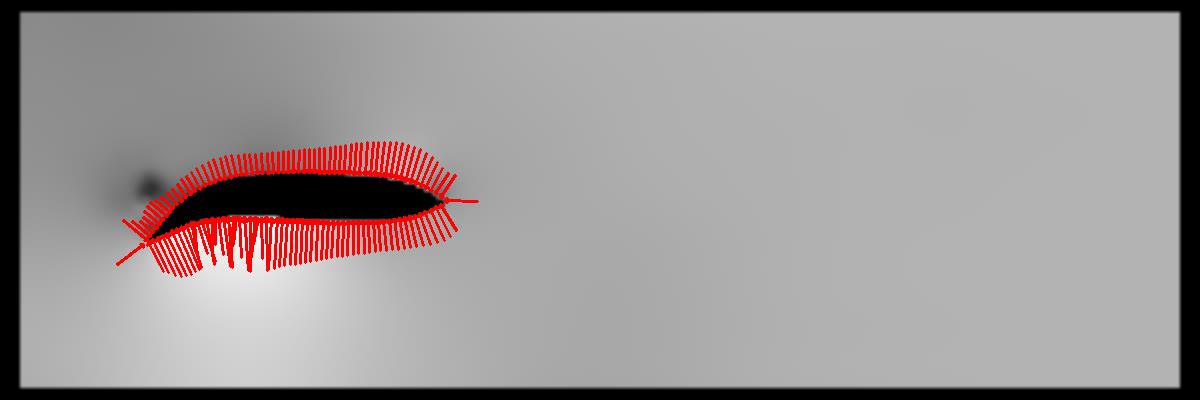} &
\includegraphics[trim=50 20 570 20,clip,width=56mm]{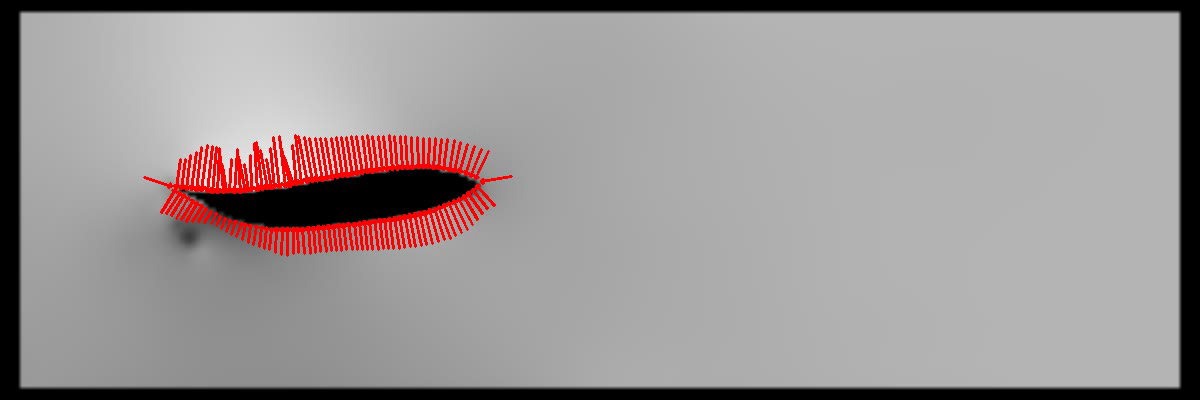} &
\includegraphics[trim=50 20 570 20,clip,width=56mm]{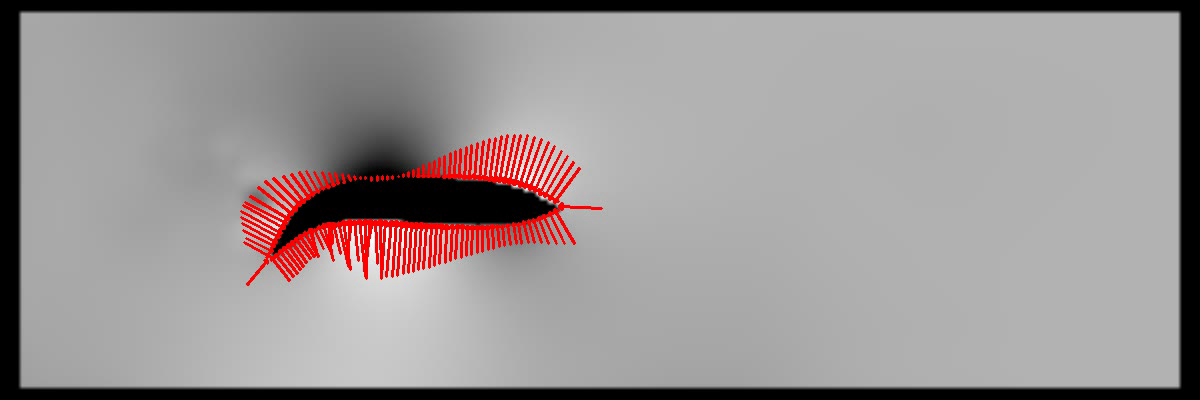}
\end{tabular}
}
\caption{The forward swimming of a carangiform soft body fish immersed in fluid is simulated using our hybrid technique featuring FEM solid simulation and neural network based hydrodynamics. The figures illustrate at three separate time steps ($t$ = 73, 162, 275) both the fish and the full fluid pressure field. The red arrows at the interface from fluid to solid indicate the forces applied from the fluid to the fish. The forces are calculated with the immersed boundary method.}
\label{fig:general}
\end{center}
\vskip -0.2in
\end{figure*}

\section{Introduction}

Soft robotics is a rapidly advancing branch of robotics, showing promising results in non-standard settings in which compliant structures and bio-inspired designs are needed to solve tasks in natural environments~\cite{hawkes_hard_2021}. Aquatic locomotion is one such setting where soft robotic designs are uniquely able to take advantage of hydrodynamic properties, mimicking biological fish designs selected through evolutionary pressures for maximum fitness in nature~\cite{katzschmann_exploration_2018}.

Simulation is a non-trivial challenge in the design of soft robots, as opposed to the rigid domain, for which established techniques have been developed and built upon in the span of decades. As for aquatic locomotion, solving fluid-structure interaction (FSI) for incompressible Navier-Stokes is a hard problem, traditionally extremely computationally expensive and thus often impossible in practice. Leveraging these simulations for design and control optimization generally involves an additional computational burden, through slow evolutionary or otherwise gradient-free optimization procedures. In this work, we leverage recent advances in machine learning for physics to take a step towards solving this problem, proposing a FSI simulation that is both orders of magnitude faster than standard approaches and fully differentiable, thus allowing for simple gradient-based optimization of design and control objectives.

To address the FSI challenge, our hybrid approach uses a differentiable numerical simulation for the deformable solid structure and a neural network surrogate to capture hydrodynamic effects of the fluid. To perform fast and differentiable soft body simulation of a flapping 2D carangiform swimmer, we leverage the finite-element method (FEM) combined with the novel approach of Differentiable Projective Dynamics (DiffPD)~\cite{du_diffpd_2021}. For the fluid simulation, we train a physics-constrained neural network for hydrodynamics as proposed by Wandel et al.~\yrcite{wandel_learning_2021}, which approximates fluidic flow on a discretized marker and cell (MAC) grid~\cite{harlow_numerical_1965}. Their approach requires no training data but is instead trained using a physics-constrained loss based on the Navier-Stokes differential equation.

\subsection{Our Contribution}

Our contributions are the following:
\begin{itemize}
    \item We introduce a differentiable layer linking the solid and the fluid simulations, achieving FSI coupling while maintaining low computational cost and full automatic differentiability. This approach involves the specification of the fluid boundary condition as a \textit{soft mask} with techniques from differentiable rendering~\cite{liu_soft_2019}, and the computation of fluid-to-solid surface forces using a variation of the Immersed Boundary Method (IBM)~\cite{peskin_immersed_2002} with Gaussian distance.
    \item We demonstrate on a 2D carangiform swimmer that our hybrid approach leads to realistic swimming behavior with forward propulsion, while requiring considerably less computational resources than existing FSI simulations (\cref{fig:general}).
    \item We leverage the differentiability of the simulation to directly optimize the frequency parameter of a swimming controller with first order gradient optimization to achieve higher forward swimming speeds.
    \item We compare our hybrid simulation with the traditional COMSOL solver for FSI, demonstrating a monotonic relationship between the distance travelled by the fish with the same controller on either simulation.
\end{itemize}

\section{Related Work}

\subsection{Soft Robotics and Aquatic Locomotion}
To date there exist several examples of successful applications of soft robots in aquatic environments. Bio-inspired artificial fish robots \cite{katzschmann_hydraulic_2016,lin_modeling_2021} mimicking biological fish have been deployed for a range of practical tasks, from exploration of underwater environments and observation of aquatic life \cite{katzschmann_exploration_2018,li_self-powered_2021} to protection of marine habitats against invasive species \cite{polverino_ecology_2022}.

Interest in the topic of aquatic locomotion has not been limited to robotics: the problem of learning swimming controls in simulation has received considerable attention from a reinforcement learning perspective \cite{gazzola_learning_2016,colabrese_flow_2017,verma_efficient_2018}, albeit relying on simple swimmer models that avoid the issues presented by soft bodies and FSI. For this reason, translating learnt controllers from these simulated works back to real-world robotics platforms (sim2real) remains an open challenge. Other works on aquatic locomotion simulation \cite{liu_computer_2015} do not involve learning controls altogether, but are limited to investigations into properties of biological swimmers through simulation.

\subsection{Neural Networks for Hydrodynamics}

Seminal work on Physics-Informed Neural Networks (PINNs) \cite{raissi_physics-informed_2019} laid the foundations for the application of machine learning models as surrogate differential equations solvers. PINN training leverages the autodifferentiation properties of neural networks to learn the dynamics of differential equation systems without the need for costly training data, but using differential equation residuals directly as the model's loss function.

A trove of subsequent works build on top of the PINN approach, refining the technique specifically for hydrodynamics problems \cite{mao_physics-informed_2020, raissi_hidden_2020, sun_surrogate_2020}. Wandel et al. \yrcite{wandel_learning_2021} propose a neural network model for hydrodynamics based on a discretized MAC grid and trained with a Navier-Stokes residual loss. While technically not following the original PINN specification, as the derivative terms in the Navier-Stokes loss are computed with finite differences on the grid as opposed to using autodifferentiation of the model inputs, their model allows for dynamic interactive specification of boundary conditions. For this reason, we adopt their model as a core component of our hybrid simulator. Further work from the same authors generalizes the technique to 3D simulations \cite{wandel_teaching_2021} and to higher turbulence fluids \cite{wandel_spline-pinn_2021}.

\subsection{Differentiable Soft-Body Simulation}

Differentiable soft-body simulation extends standard soft-body simulators to compute gradients for a soft body's shape, control, or state parameters. Such gradients have been proven helpful in several downstream soft robotic applications, including system identification~\cite{hu_chainqueen_2019}, trajectory optimization~\cite{geilinger_add_2020}, motion control~\cite{qiao_differentiable_2021}, and shape optimization~\cite{ma_diffaqua_2021}.

Mainstream differentiable soft-body simulators fall into two categories: physics-based and learning-based. Physics-based differentiable soft-body simulators derive gradients based on governing equations characterizing system dynamics and require domain-specific knowledge~\cite{hu_chainqueen_2019, geilinger_add_2020, hu_difftaichi_2020, du_diffpd_2021, qiao_differentiable_2021}. On the other hand, learning-based approaches aim to learn a neural network model approximating soft-body dynamics~\cite{li_learning_2019,pfaff_learning_2021}. Such neural networks are naturally differentiable but, unlike their physics-based counterparts, typically lack guarantees on physics invariants, \emph{e.g.}, energy or momentum conservation. Additionally, their generalizability to new settings largely depends on the quality of the training data.

In this work, we use DiffPD \cite{du_diffpd_2021}, a recently developed physics-based differentiable soft-body simulator, to simulate our aquatic swimmers. While our algorithm is agnostic to the choice of differentiable simulators, we have found DiffPD beneficial because of its faster speed compared to other physics-based differentiable soft-body simulators and its extension to underwater robotics applications~\cite{ma_diffaqua_2021,du_underwater_2021}.

\subsection{Fluid-Structure Interactions}
Fluid-structure interaction (FSI) studies the complex behavior of fluids coupled with solid objects in a multi-physics system. FSI has been an extensively studied research topic in mechanical engineering, computational physics, and other related fields for many years~\cite{dowell_modeling_2001}. Existing works on FSI typically focus on coupling fluids and rigid or soft objects under small deformations \cite{mucha_model_2004,zhang_two_2007,kalitzin_toward_2003,yang_embedded-boundary_2006}. Numerical techniques for coupling fluids and nonlinear soft objects with large deformation have also been developed in computational physics and computer graphics \cite{feng_simulations_2019,brandt_reduced_2019,robinson-mosher_two-way_2008,lu_two-way_2016,teng_eulerian_2016,fang_iq-mpm_2020}. Existing methods for FSI have been used to investigate the swimming behavior of fish \cite{curatolo_virtual_2015}. These techniques provide accurate yet expensive computational tools for applications such as underwater soft swimmers covered in this work.

None of the FSI methods described above take into account gradient computation. Because of the already complicated coupling between fluids and solids, works on \emph{differentiable} FSI are unsurprisingly sparse. Existing differentiable FSI approaches typically rely on non-physical simplifications of fluid, solid, or coupling models, \emph{e.g.}, assuming simple empirical fluid models~\cite{ma_diffaqua_2021,du_underwater_2021} or limiting interactions to be local~\cite{li_learning_2019,pfaff_learning_2021}. Compared with those previous works, our work is different because we attempt to incorporate the full physics in all possible aspects: Navier-Stokes equations for modeling fluids, continuum mechanics for modeling soft bodies, and immersed boundary methods \cite{peskin_immersed_2002} (IBM), a representative FSI method, for modeling solid-fluid coupling.

\section{Differentiable FSI Method}

\begin{figure}[ht]
    \begin{center}
    \centerline{\includegraphics[width=\columnwidth]{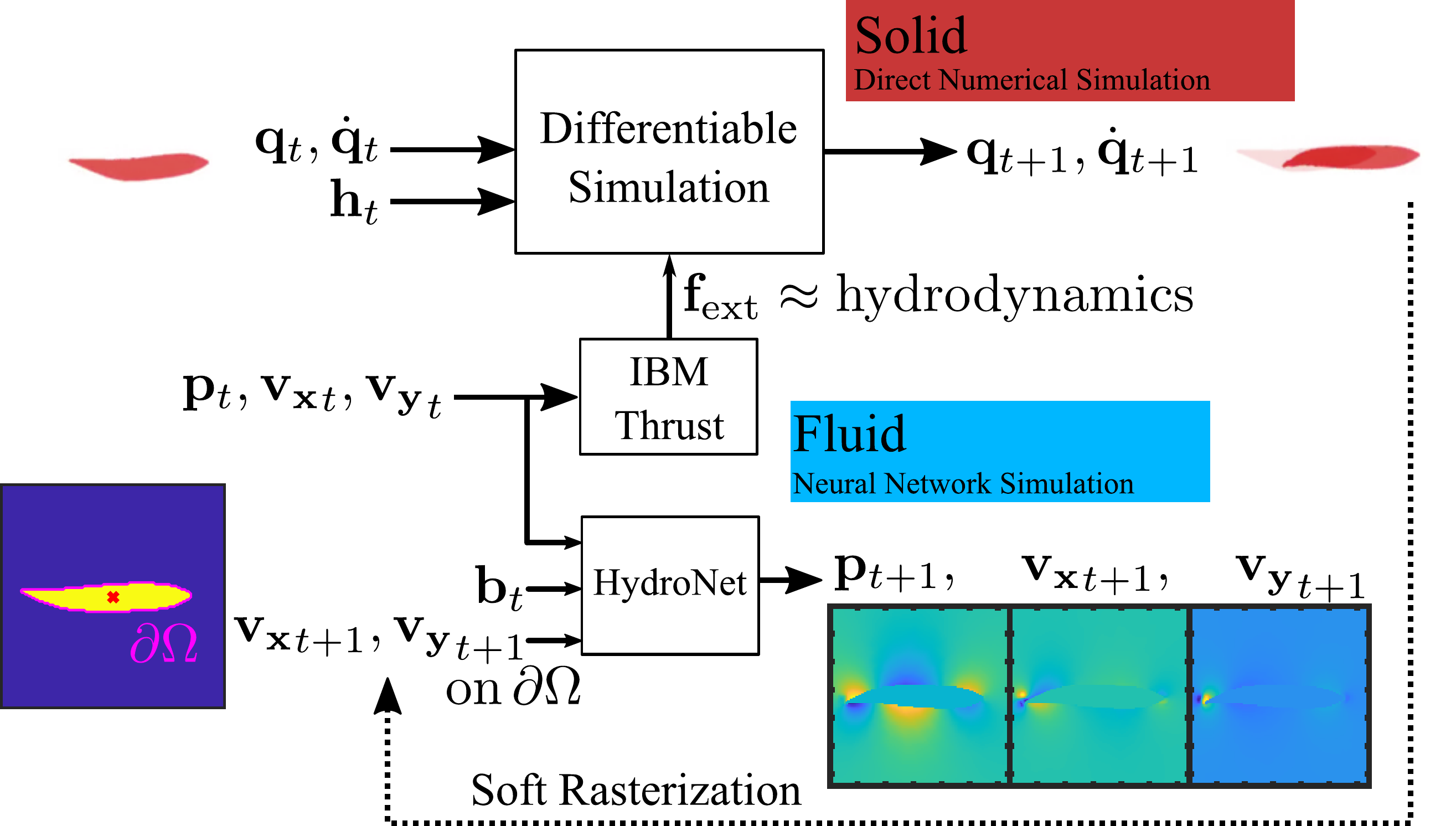}}
    \caption{Overview block diagram of our hybrid simulation method. $\vq$, $\dot{\vq}$ are positions and velocities of finite elements; $\vh$ are actuator signals; $\vp$, $\vv_x$ and $\vv_y$ are pressure and velocity fields of the fluid; $\vb$ is the soft boundary mask; $\mathbf{f}_{\mathrm{ext}}$ is the hydrodynamic force applied by the fluid to the solid.}
    \label{fig:overview}
    \end{center}
    \vskip -0.2in
\end{figure}

We hereby detail our hybrid method for fast and fully differentiable simulation of soft body and fluid interaction. As shown in Figure~\ref{fig:overview}, our approach consists of repeated, stacked interaction between the DiffPD solid simulator and the Hydrodynamics Neural Network (HydroNet) surrogate simulator. The output of the fluid simulation at time $t$ is used as input for the solid simulation at the same time $t$ through the introduction of an external fluidic force, while the output of the solid simulation at time $t$ is used as input for the fluid simulation at the next time step $t+1$ through the specification of a boundary condition. This interleaved interaction leads to the unfolding of a differentiable computation graph that can be backtraced by autodifferentiation frameworks to compute gradients through the entire simulation episode and optimize objectives (\cref{fig:sim-graph}). FEM element positions and velocities, actuations, fluid curls and pressures, boundary conditions, external forces and Young's moduli are all differentiable with respect to each other. In the following sections, we describe each component of the system in detail.

\subsection{Hydrodynamic Surrogate Simulation}

Our goal for fluid simulation is that of obtaining a fast and differentiable surrogate model of hydrodynamics, so that swimmer designs and controls can be optimized by differentiating directly against the simulation environment, as opposed to doing so through evolutionary strategies or reinforcement learning.

In particular, our surrogate simulator needs to solve the incompressible Navier-Stokes equations describing fluidic flow costrained by Dirichlet boundary conditions. Given fluid density $\rho$ and viscosity $\mu$, and defining $\mathbf{v}$ and $\mathbf{p}$ to be velocity and pressure fields over a fluid domain $\Omega$, the equations consist of the following three terms:
\begin{align}
    \nabla \cdot \mathbf{v} = 0 &\qquad \text{on }\Omega\text{,}\label{nv-term1}\\
    \rho \left(\frac{\partial \mathbf{v}}{\partial t} + (\mathbf{v} \cdot \nabla) \mathbf{v} \right) = - \nabla \mathbf{p} + \mu \nabla^2 \mathbf{v} &\qquad \text{on }\Omega\text{,}\label{nv-term2}\\
    \mathbf{v} = \mathbf{v}_d &\qquad \text{on }\partial\Omega\text{.}\label{nv-term3}
\end{align}
Equation (\ref{nv-term1}) is called the divergence term, and forces incompressibility of the fluid, disallowing any sources or sinks within $\Omega$. Equation (\ref{nv-term2}) is the main hydrodynamic term, stating that changes in fluidic particle momentum must correspond to forces exerted by the pressure gradient and viscous friction. Equation (\ref{nv-term3}) is the Dirichlet boundary condition, stating that velocities on the boundary $\partial\Omega$ of the fluidic domain must be equal to the supplied boundary velocities $\mathbf{v}_d$.

The equations can be simplified for the 2D case by observing that the Helmholtz theorem allows to decompose the velocities into a curl free part and a divergence-free part
\begin{equation}
\mathbf{v} = \nabla\mathbf{q} + \nabla \times \mathbf{a}\text{.}
\end{equation}
By expressing velocities as $\mathbf{v} = \nabla \times \mathbf{a}$, with the curl $\mathbf{a}$ being one-dimensional in the 2D case, we implicitly force zero-divergence in our fluid without the need for explicitly solving for Equation (\ref{nv-term1}).

To solve the equations, we train an unsupervised neural network model following the approach proposed by Wandel et al. \yrcite{wandel_learning_2021}. The model in focus is a U-net \cite{ronneberger_u-net_2015} with limited convolutional channels that operates on discretized fields on a marker and cell (MAC) \cite{harlow_numerical_1965} grid. The model takes as input the curl field $\mathbf{a}$, the pressure field $\mathbf{p}$, the boundary mask $\mathbf{b}$ identifying the domain $\Omega$, and the boundary velocities $\mathbf{v}_d$ from time step $t$. The model predicts as output the curl field $\mathbf{a}$ and the pressure field $\mathbf{p}$ for the next time step $t+1$, given a constant time step magnitude $h$.

The model is not trained on simulation data, but instead uses the Navier-Stokes residuals as its loss function:
\begin{align}
    L_p &= \left\|\rho \left(\frac{\partial \mathbf{v}}{\partial t} + (\mathbf{v} \cdot \nabla) \mathbf{v} \right) + \nabla \mathbf{p} - \mu \nabla^2 \mathbf{v} \right\|^2\quad \text{on }\Omega\text{,}\label{nvloss-net}\\
    L_b &= \left\|\mathbf{v} - \mathbf{v}_d \right\|^2 \quad \text{on }\partial\Omega\text{,}\label{boundloss-net}\\
    L &= \beta L_p + \gamma L_b \text{.}\label{totloss-net}
\end{align}
with parameters $\beta$ and $\gamma$ determining how much to prioritize the Navier-Stokes term or the boundary term.

Given this loss, the network is trained on synthetic episodes consisting of randomly generated boundary conditions, with the network's curl and pressure fields output being fed back as training data at each iteration. This way, the physics-constrained loss is applied to increasingly realistic scenarios.




\subsection{Differentiable Soft-Body Simulation}

We model the soft-body dynamics by the following governing equations from continuum mechanics~\cite{sifakis_fem_2012}:
\begin{align}
    \rho_s \ddot{\vq} = \nabla \cdot \mathbf{P} + \mathbf{f}_{\textrm{ext}},
\end{align}
where $\rho_s$ stands for the soft material's density, $\vq = \vq(\mathbf{X}, t)$ tracks the position of a material point $\mathbf{X}$ from the material space (undeformed shape) at time $t$, $\mathbf{P}$ represents the first Piola-Kirchoff stress tensor, and $\mathbf{f}_{\text{ext}}$ captures all external forces applied to $\mathbf{X}$ at time $t$. We refer interested readers to Sifakis \& Barbic \yrcite{sifakis_fem_2012} for more background information regarding soft-body simulation. The stress $\mathbf{P}$ determines the behavior of soft material and is specified by choosing soft material models. In this work, we use the linear co-rotated material model as suggested by DiffPD because of its balance between speed~\cite{bouaziz_projective_2014} and physical accuracy~\cite{du_underwater_2021,zhang_learning_2021}.

Given the continuous equations above, DiffPD uses standard finite-element methods (FEM) and the implicit time-stepping scheme to discretize the dynamic system spatially and temporally, leading to the nonlinear system of equations below:
\begin{align}
\vq_{t + 1} &= \vq_t + h \dot{\vq}_{t + 1}, \\
\dot{\vq}_{t + 1} &= \dot{\vq}_t + h \rho_s^{-1}(\mathbf{f}_{\textrm{ext}} + \mathbf{f}_{\textrm{int}}(\mathbf{x}_{t + 1})),
\end{align}
where $\vq$ and $\dot{\vq}$ now represent nodal positions and velocities of finite elements at time steps specified by their subscripts. The notation $\mathbf{f}_{\text{int}}$ denotes the elastic force induced by the stress tensor $\mathbf{P}$. Once the forward simulation process is established, DiffPD derives its gradients using standard chain rules and adjoint methods. Interested readers can refer to DiffPD~\cite{du_diffpd_2021} for detailed derivations of these equations.




\subsection{Differentiable Fluid-Structure Interaction}

FSI involves solving a two-way link between the DiffPD soft structure FEM simulation and our neural network hydrodynamic surrogate simulation. Hydrodynamic forces from the fluid simulation affect the soft body finite elements as an external force $\mathbf{f}_\textrm{ext}$, at the same time the soft body simulation determines the Dirichlet boundary conditions $\mathbf{b}$ and $\mathbf{v}_d$ for the hydrodynamics simulation.
An additional cause of complexity stems from the fact that these operations must mediate between a \textit{Lagrangian} and a \textit{discrete Eulerian} representation for physical quantities, with the former being used the the solid simulation, and the latter for the fluid simulation.

Lagrangian methods handle physics simulation by modeling individual particles constituting the simulated material. DiffPD operates in a Lagrangian fashion, as the finite elements identified by $\vq$ and $\dot{\vq}$ track specific points within the soft body and move along with the body within the domain. Opposed to this, discrete Eulerian methods simulate PDEs on a discretized grid such as the fixed MAC grid used by Wandel et al. \yrcite{wandel_learning_2021}'s hydrodynamics network. With this representation, $\mathbf{v}$ and $\mathbf{p}$ summarise fluid properties on fixed locations of the domain, without tracking individual fluid particles.

The challenge in this setting is that of providing a differentiable layer to compute these interaction quantities which mediate between representations. For the solid-to-fluid interaction, the Lagrangian elements described by $\vq$ and $\dot{\vq}$ must be used to compute a boundary mask $\mathbf{b}$ with a rasterization operation, which is however generally non-differentiable. Similarly, grid boundary velocities $\mathbf{v}_d$ are generally computed with a non-differentiable neighbourhood averaging operation. For the fluid-to-solid interaction, we turn to the fluid-to-solid stage of the Immersed Boundary Method (IBM) \cite{peskin_immersed_2002}, which samples Eulerian pressure values $\mathbf{p}$ on locations near the boundary in order to compute Lagrangian external forces $\mathbf{f}_\textrm{ext}$ affecting the solid finite elements.

\paragraph{Solid-to-Fluid Coupling}

Given the state of the DiffPD soft body simulation from a specific time step, the finite element positions $\vq$ and velocities $\dot{\vq}$ fully determine the boundary condition for the subsequent hydrodynamic simulation step. The positions $\vq$ determine the shape and location of the fish body, served as input to the Hydronet as a boundary mask $\mathbf{b}$. The velocities $\dot{\vq}$ are instead used to compute the boundary velocities $\mathbf{v}_d$ as a granular cell-wise velocity obtained by averaging the closest elements to each boundary cell.

There is however a non trivial obstacle that renders a naive application of the boundary mask from Wandel et al.~\yrcite{wandel_learning_2021} non applicable to our optimization setting. By definition, the rasterization operation computing a binary mask is non-differentiable, thereby breaking the chain of differentiability which we rely on for optimization.

Our solution to this issue takes from the field of differentiable rendering \citep{liu_soft_2019}, and in particular to the techniques associated with soft rasterization. Instead of producing a hard binary mask as a rasterization of the robot's finite element mesh, we use a signed distance field to produce a soft differentiable mask, with real-valued entries $\mathbf{b}_{ij} \in [0,1]$.

Define $\vx_{ij} \in \mathbb{R}^2$ as the spatial coordinates of the MAC grid cell in position $(i,j)$ and $\vq^k$ those of the $k$-th finite element. Then each cell $\vb_{ij}$ from the soft boundary mask is computed as
\begin{equation}
    \mathbf{b}_{ij} = \text{sigm.}\left( \delta_{ij} \frac{\sum_{l} \left( \text{softmin}_{k}\frac{\|\vx_{ij}- \vq^k\|^2}{\xi}\right)_l \|\vx_{ij}- \vq^l\|^2}{\sigma} \right)
\end{equation}
with $\sigma$ and $\xi$ being mask softness parameters,  $\|\cdot\|^2$ the Euclidean distance and $\delta_{ij} = +1$ if the cell location $\vx_{ij}$ is inside the fish body, while $\delta_{ij} = -1$ if outside.

We can similarly use a differentiable surrogate to obtain cell-wise fine-grained boundary velocities:
\begin{equation}
    {\vv_d}_{ij} = \left( \sum_{l}{ \left( \text{softmin}_{k} \frac{\|\vx_{ij}- \vq^k\|^2}{\tau}\right)_l {\dot{\vq}}^l} \right) \cdot \vb_{ij}
\end{equation}
with $\tau$ being a softness parameter.

Tuning the softness parameters $\sigma$, $\xi$ and $\tau$ allows us to tune the trade-off between boundary accuracy and smooth gradients.

\begin{figure}[ht]
\vskip 0.2in
\begin{center}
\centerline{\includegraphics[trim=10 40 25 40, clip,width=0.8\columnwidth]{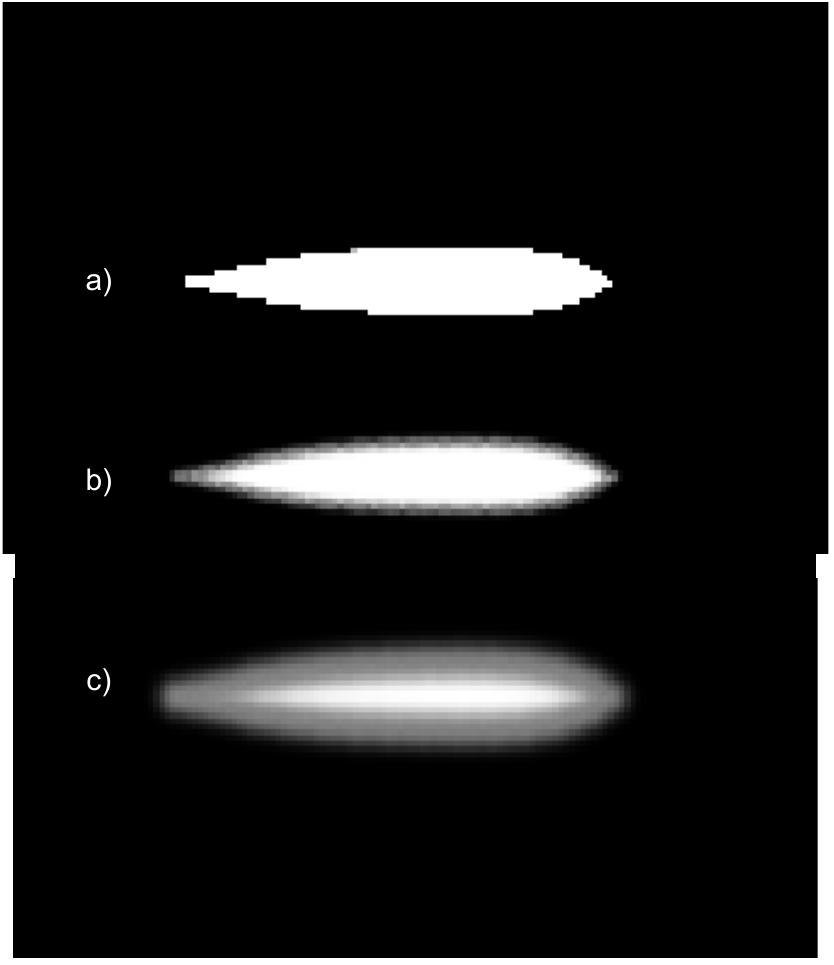}}
\caption{Soft boundary masks obtained with the softness parameters $\xi=$ 5e-7 and a) $\sigma =$ 5e-9, b) $\sigma =$ 5e-7, c) $\sigma =$ 5e-5.}
\label{fig:soft-boundary-mask}
\end{center}
\vskip -0.2in
\end{figure}

\paragraph{Fluid-to-solid coupling}

The overall purpose of our hybrid simulation approach is to optimize fish designs and/or control policies with respect to a more realistic model of hydrodynamics. Therefore, the mechanism by which the hydrodynamics simulation affects the soft body simulation is of utmost importance.

The way for hydrodynamic forces to affect DiffPD is through an external force $\mathbf{f}_\textrm{ext}$ applied to the simulation's finite elements. Common drag/thrust optimizations such as that of Chen~et~al.~\yrcite{chen_numerical_2021} often average the force over the entire solid surface. This force is shaped by two contributions, one due to the pressure field,
\begin{equation}
    \mathbf{f}_\textrm{pressure} = - \int_{\partial \Omega} \mathbf{p} \cdot \mathbf{n} \,dl,
\end{equation}
and the other is due to the velocity field,
\begin{equation}
    \mathbf{f}_\textrm{viscous} = - \int_{\partial \Omega} \mu \mathbf{n} \times \mathbf{a} \,dl \text{,} \quad \mathbf{a} = \nabla \times \mathbf{v},
\end{equation}
with $\partial \Omega$ being the solid body boundary and $\mathbf{n}$ its outward pointing normal vector. Total hydrodynamic force is obtained by summing $\mathbf{f}_\textrm{ext} = \mathbf{f}_\textrm{pressure} + \mathbf{f}_\textrm{viscous}$. However, for common water-like fluids with low viscosity $\mu$, the contribution of the viscous term $\mathbf{f}_\textrm{viscous}$ can be considered negligible and its computation can be omitted.

\begin{figure}[ht]
\vskip 0.2in
\begin{center}
\centerline{\includegraphics[trim= 157 336 221 0, clip, width=0.8\columnwidth]{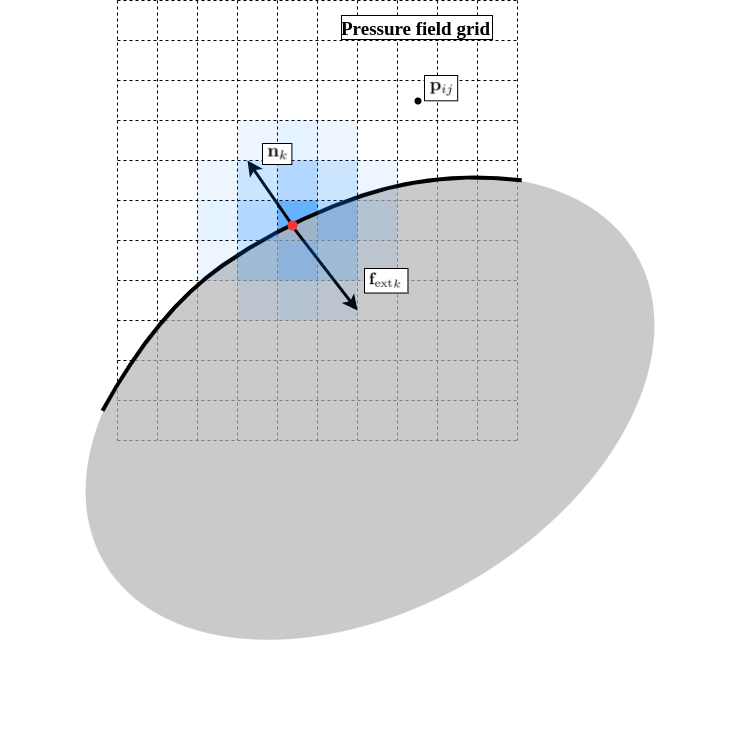}}
\caption{Our immersed boundary method for fluid-to-solid interaction. Each soft body surface element (red) is subjected to an external force ${\mathbf{f}_{\mathrm{ext}}}_k$ in the opposite direction of its normal $\mathbf{n}_k$. To compute the scalar force magnitude, nearby Eulerian cells from the pressure field are averaged with a Gaussian function centered around the element.}
\label{fig:ibm}
\end{center}
\vskip -0.2in
\end{figure}

Given that our approach for solid simulation is based on finite elements, with each surface element $k$ being associated with its surface normal $\vn_k$, we can compute individual elements' surface forces ${\mathbf{f}_\textrm{ext}}_k$. To bridge the Eulerian to Lagrangian gap, we adopt the IBM fluid-to-solid step on forces
\begin{align}
    {\mathbf{f}_\textrm{ext}}_k = - l_k \vn_k \sum_{i,j}{ \vp_{ij} \delta\left( \mathbf{x}_{ij} - \vq^k \right) \vb_{ij}}
\end{align}
where $\delta$ is the Dirac delta and $l_k = (\|\vq^{k-1} - \vq^{k}\|+\|\vq^k - \vq^{k+1} \|)/2$ is the surface length corresponding to finite element $k$. With the IBM, we are able to appropriately identify the force applied to Lagrangian element $k$, despite only having access to pressures on a fixed Eulerian grid.

In practice, due to the finite discretization, it is not feasible to adopt $\delta$ directly, but a surrogate $\tilde{\delta}(\vx) = \phi(x_1)\phi(x_2)$ must be chosen such that it satisfies several properties as detailed in the original IBM paper \cite{peskin_immersed_2002}:
\begin{align}
    \phi(r) \text{ is continuous for all real }r\text{,}\\
    \sum_{j < r} \phi(r-j) = \sum_{j > r} \phi(r-j) = \frac{1}{2} \text{ for all real }r\text{,}\\
    \sum_{j} (r-j) \phi(r-j) = 0 \text{ for all real }r\text{,}\\
    \sum_{j} (\phi(r-j))^2 = C \text{ for all real }r\text{,}
\end{align}
where the constant $C$ is independent of $r$. The original formulation from Peskin \yrcite{peskin_immersed_2002} included the additional property of $\phi(r) = 0$ for $|r|>2$, however this is not strictly required and is only introduced for computational cost reasons, which are moot if the operation is performed with GPU parallelism.

We thus choose to use a normalized Gaussian distance for our IBM, namely in the form
\begin{align}
    \tilde{\delta}(\vx - \vy) = \exp\left(-\frac{\|\vx - \vy\|^2}{2\sigma'} \right) \text{,}
\end{align}
with $\sigma'$ being a smoothness parameter and the equation satisfying all relevant properties.

This choice then leads to our IBM formula for calculating individual elements' ${\mathbf{f}_\textrm{ext}}_k$:
\begin{align}
    {\mathbf{f}_\textrm{ext}}_k = - l_k\vn_k \sum_{i,j}{ \vp_{ij} \frac{1}{Z}\exp\left(-\frac{\|\mathbf{x}_{ij} - \vq^k\|^2}{2\sigma'} \right) \vb_{ij}} \text{,}
\end{align}
with $Z=\sum_{i,j} \exp\left(-\frac{\|\mathbf{x}_{ij} - \vq^k\|^2}{2\sigma'} \right) \vb_{ij}$ being a normalization constant.

The reason we choose a Gaussian delta function is not only because of the increased stability due to larger function support (as is discussed by Peskin~\yrcite{peskin_immersed_2002}). Using a Gaussian as opposed to the original Dirac delta gives us the property of differentiability of the forces with respect to the entire pressure field. Once again, there is a trade-off between gradient smoothness and IBM precision, as lower $\sigma'$ allows for more precise IBM interpolation, but causes the gradients with respect to the pressure field to vanish for most locations.

The obtained $\mathbf{f}_{\mathrm{ext}}$ is at the granularity of individual finite elements. It can either be applied as a DiffPD input as-is, allowing for precise but potentially unstable simulation of surface interaction, or it can be used to compute the overall thrust/drag
\begin{align}
    \mathbf{f}_\textrm{ext} = \sum_k {\mathbf{f}_\textrm{ext}}_k \label{averaged-f}
\end{align}
which, divided by the total finite element number, can be applied as an average force to all the solid finite elements, to only model directional thrust.

\subsection{Limitations}

Our FSI simulator is fast and differentiable, and its hydrodynamics component is trainable on several different fluid parameter settings, also supporting both still and moving flow scenarios for swimmers (through setting of the inflow velocity as a boundary condition). However, this does not mean that the method is without limitations. For instance, the method does not generalize to flow velocities well outside those imposed by inflow boundary conditions during HydroNet training. The simulation can also present instabilities if the forces involved are too strong or applied suddenly between frames: for this reason we explictly set a \textit{burn-in} number $N_{\text{burn-in}}$ of iterations during which surface forces $\mathbf{f}_\textrm{ext}$ are linearly smoothed by a factor of $t/N_{\text{burn-in}}$.

\section{Experimental Setup and Results}

\subsection{Soft Carangiform Swimmer}\label{experimental-carangiform}

Our main experimental setting is controller optimization for a soft body carangiform swimmer immersed in Navier-Stokes fluid with parameters resulting in various degrees of turbulence.

\begin{figure}[ht]
\vskip 0.2in
\begin{center}
\centerline{\includegraphics[width=0.7\columnwidth]{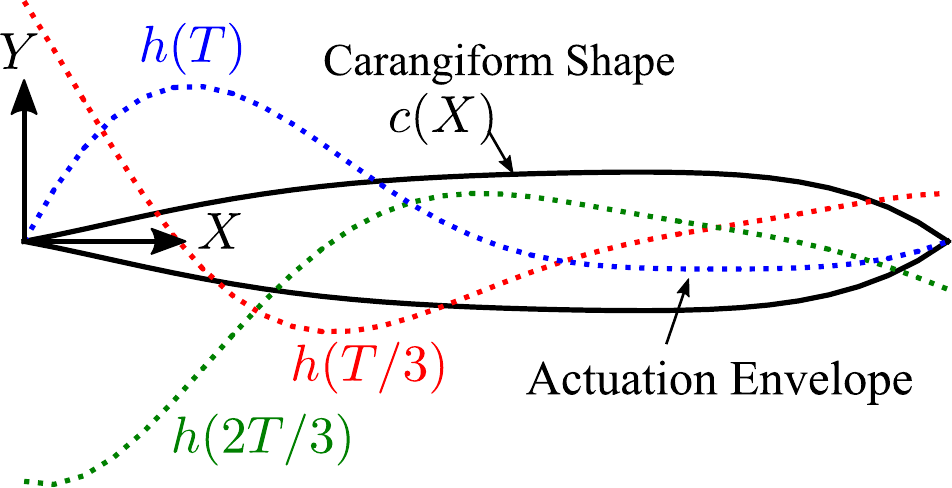}}
\caption{Coordinate system of the fish in the material frame, where $c(X)$ is the parametric shape of the carangiform fish and $h(X,t)$ is the actuation envelope (defined in Appendix \ref{appendix:fish}).}
\label{fig:carangiform}
\end{center}
\vskip -0.2in
\end{figure}

The swimmer's profile is generated using a parametric polynomial $c(X)$ adapted from Curatolo \& Teresi \yrcite{curatolo_virtual_2015} and the actuation envelope is obtained with a parametric curve adapted from Videler \& Hess \yrcite{videler_fast_1984} as shown in Figure \ref{fig:carangiform}. This parametric shape is discretized as a mesh, which is used to define the finite elements for DiffPD. We set the fish body's Youngs' modulus as $E = \SI{6e5}{\pascal}$, its Poisson ratio as $\nu = 0.45$ and its density as $\rho_{s} = \SI{100}{\kilogram\per\metre^3}$. We refer to Appendix \ref{appendix:fish} for further details.

\subsection{HydroNet Training}

We model our fluid medium as a quadrilateral 2D box of size $\SI{0.25}{\metre} \times \SI{0.75}{\metre}$, discretized as a $100 \times 300$ MAC grid with resolution of $\SI{2.5}{\milli\metre}$, and computed with time step $h = \SI{0.01}{\second}$.

To immerse the solid simulation in a suitable water-like environment at this scale, we retrain the hydrodynamics network from Wandel et al.~\yrcite{wandel_learning_2021}, performing a hyperparameter search on model hyperparameters and loss scalings to train for density $\rho \in [ \SI{10}{\kilogram\per\metre^3},  \SI{50}{\kilogram\per\metre^3}]$ and viscosity $\mu \in [ \SI{0.125}{\milli\pascal\cdot\second}, \SI{1}{\milli\pascal\cdot\second} ]$. The resulting hydrodynamic networks model a variety of fluid environments, from conditions close to laminar flow to turbulent, light-water scenarios.

The main tuning challenge for training the network on the desired fluid parameters is that of re-scaling the terms appearing in the losses (Equations (\ref{nvloss-net}) and (\ref{boundloss-net})), which in Wandel et al.~\yrcite{wandel_learning_2021} operate on time and space resolutions of whole seconds and metres. Given that we require a much finer resolution, re-scaling the equation terms by a factor of $1/0.0025 = 400$ allowed the training to converge.

\subsection{FEM Validation with COMSOL}\label{experiment:comsol}

Our hybrid simulator has the properties of being fast and differentiable. However, the HydroNet component of our simulator makes it fundamentally a surrogate model with no strict numerical guarantees. It is therefore apparent that while using such a hybrid simulator model for design and control optimization will result in much faster convergence towards a solution, it will not evaluate such solution with respect to the underlying physical ground truth.

\begin{figure}[ht]
\begin{center}
\centerline{\includegraphics[width=0.95\columnwidth]{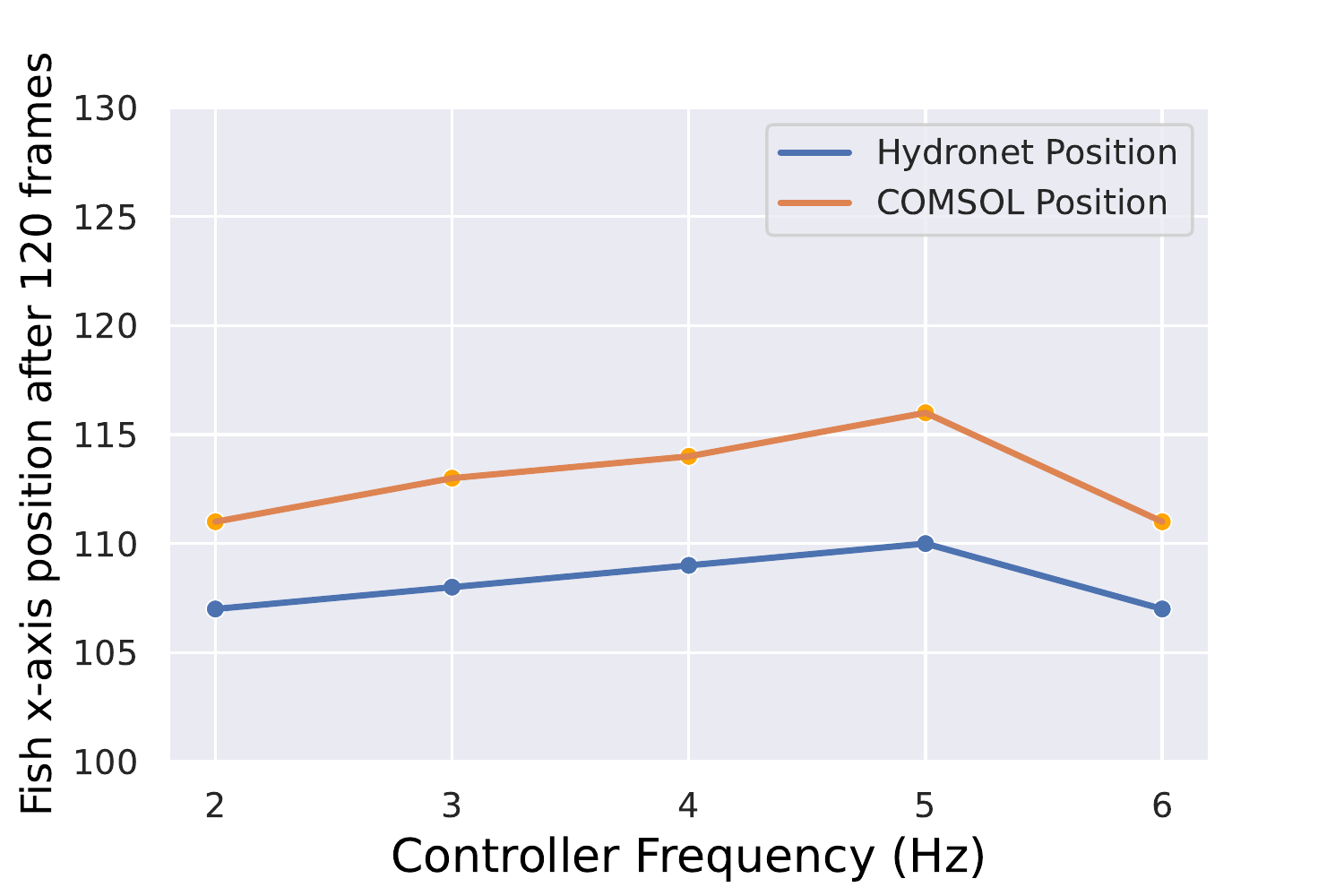}}
\caption{Comparison of travelled distance on the x-axis between COMSOL Multiphysics and our HydroNet simulator.}
\label{fig:comsol-comp}
\end{center}
\vskip -0.1in
\end{figure}

We therefore propose to benchmark our simulator against \emph{COMSOL Multiphysics} for the forward swimming carangiform task, using FEM numerical simulation to model both hydrodynamics, soft body physics, and FSI (see Appendix~\ref{appendix:comsol} for further details). Our comparison of a surrogate model against a slow, non-differentiable, but physically-validated simulation is akin to performing sim2real validation (more appropriately, sim2sim). For a further baseline, we attempted a comparison with a heuristic hydrodynamic formula from the work of Min et al.~\yrcite{min_softcon_2019} commonly used in applications, however, this produced overly non-physical results, resulting in wholly incomparable motion.

Due to inherent chaotic behavior of the hydrodynamic system, it is impossible to perform direct step-by-step comparisons between simulation quantities, as any small difference can cause explicit residuals between the simulations to diverge. We therefore instead compare overall fish traveled distance between the simulations for different frequency parameter values of the controller: if the simulations maintain monotonicity of performance as a function of the parameter, then optimizing the parameter on the surrogate simulation will achieve high performance on the costly, realistic simulation. This is in contrast to Min et al.~\yrcite{min_softcon_2019}'s baseline, which results in off-scale, incomparable motion and thus would offer no guarantees on physical optimality.

\begin{table}[ht]
\vskip -0.1in
\caption{Runtime comparison between our \emph{hybrid simulator} and \emph{COMSOL Multiphysics} for a 300 frame episode. Note that COMSOL Multiphysics does not produce gradients and a fair comparison is therefore only to be made between COMSOL Multiphysics's total runtime and our simulator's forward pass. Also note that due to gradient checkpointing, the backward pass for our approach is slower than theoretically achievable.}
\label{tab:runtime-comsol}
\begin{center}
\begin{small}
\begin{sc}
\begin{tabular}{lr}
\toprule
Hybrid sim (ours) & Time \\
\midrule
HydroNet Warmup    & $\SI{44}{\second} \pm \SI{0.2}{\second}$\\
Total Forward pass  &   $\SI{3}{\minute}$ $\SI{33}{\second} \pm \SI{16.8}{\second}$\\
\hspace{2mm} Forward pass: DiffPD & $\SI{2}{\minute}$ $\SI{50}{\second} \pm \SI{16.5}{\second}$\\
\hspace{2mm} Forward pass: Hydronet & $\SI{43}{\second} \pm \SI{0.3}{\second}$\\
Backward pass   &   $\SI{3}{\minute}$ $\SI{57}{\second}\pm \SI{17.9}{\second}$\\
\toprule
COMSOL Multiphysics  &   Time \\
\midrule
Total runtime    &   $\SI{5}{\hour}$ $\SI{16}{\minute}$ $\SI{39}{\second}$\\
\bottomrule
\end{tabular}
\end{sc}
\end{small}
\end{center}
\vskip -0.05in
\end{table}

We compare our carangiform swimmer simulation against COMSOL Multiphysics, modeling the swimmer as described in Section~\ref{experimental-carangiform} immersed in fluid with $\rho = \SI{50}{\kilogram\per\metre^3}$ and $\mu = \SI{0.125}{\milli\pascal\cdot\second}$. Figure~\ref{fig:comsol-comp} illustrates the comparison in terms of x-axis location of the swimmer's head at the end of an episode of 120 frames ($\SI{1.2}{\second}$) for controllers with frequency between 3 and $\SI{7}{\hertz}$. As apparent, despite the presence of a systematic mismatch of on average 6 grid-units (equal to $6 \cdot \SI{2.5}{\milli\metre}$ = $\SI{15}{\milli\metre}$), both simulations behave monotonically with respect to optimality of the frequency parameter, achieving the same maximum of travelled distance at $\SI{5}{\hertz}$. In Table~\ref{tab:runtime-comsol}, we compare runtimes for our simulator and COMSOL, observing a speedup in the order of 100x.

\begin{figure}[ht]
\begin{center}
\centerline{\includegraphics[width=\columnwidth]{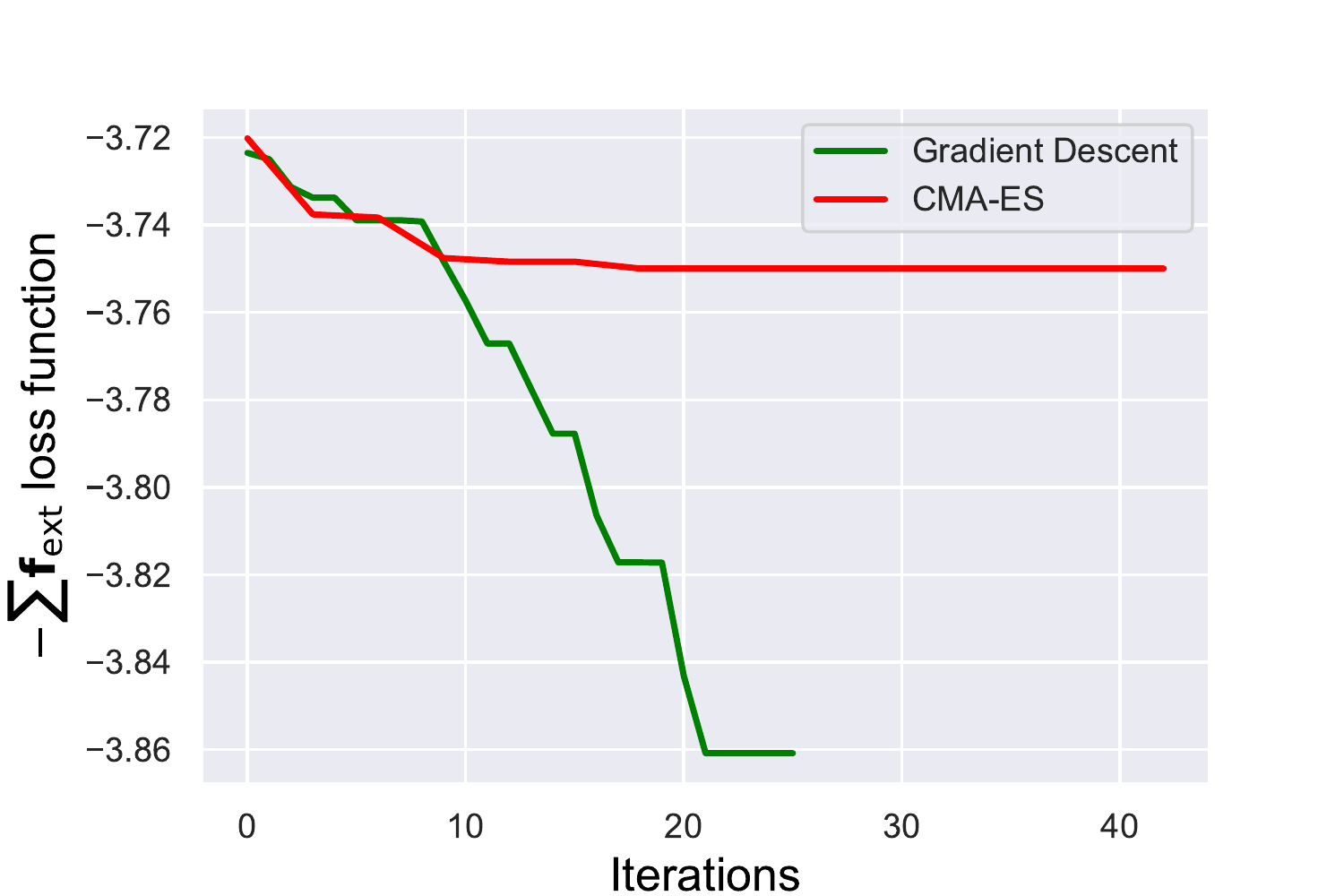}}
\caption{History plot of the controller frequency optimization, comparing gradient descent with the gradient-free CMA-ES method.}
\label{fig:optimization}
\end{center}
\vskip -0.2in
\end{figure}

\subsection{Controller Frequency Optimization}

To demonstrate the typical use case for our fully differentiable hybrid simulator, we set up a forward swimming task for the carangiform swimmer with the controller as defined in Section~\ref{experimental-carangiform}, optimizing the total forward thrust $L_{\text{opt}} = - \sum_{t \in [1,T]} \mathbf{f}_{\mathrm{ext}}(t)_x$ for a simulation episode of length $T=500$ and time step $h=\SI{0.01}{\second}$. Moreover, we fix the fish's movement to be centered along the x axis, to favor modeling of forward speed only. The objective is to optimize the controller's angular frequency $\omega$. We apply pressure thrust forces to the fish as an averaged vector $\mathbf{f}_{\mathrm{ext}}$ to limit drift and focus the optimization on swimming performance. To backpropagate over the long episode within memory limits, we resort to gradient checkpointing in between simulation steps.

The optimization experiment of the controller frequency was done in fluid medium with $\rho = \SI{50}{\kilogram\per\metre^3}$ and $\mu = \SI{0.125}{\milli\pascal\cdot\second}$. The experiment was initialized with a frequency of $\SI{2}{\hertz}$ and used Adam optimizer. As seen in Figure \ref{fig:optimization}, the optimization converged after 21 iterations to a frequency of $\SI{6.3}{\hertz}$. As benchmark comparison, the evolutionary strategy CMA-ES \cite{hansen_cma_2016}, initialized with an uninformative standard deviation of $\sigma=1$, never leaves its initialization neighbourhood within the optimization budget, obtaining its best value at the frequency of $\SI{2.63}{\hertz}$. The found optimum is different from that of Section \ref{experiment:comsol} due to our fixing of the x axis, ignoring any sideways motion.

\section{Conclusion}

In this work, we introduced a fast, differentiable and accurate FSI simulation for immersed soft bodies. We demonstrated the suitability of our FSI simulation for simulating forward swimming motion of a carangiform swimmer, and we tested its use for the optimization of the frequency of a swimming controller. We believe our method paves the way for more complex but fast multi-physics simulations that couple fluid mechanics with continuum mechanics in a unified framework. An extension of our method has the potential to in the future enable ambitious works on physically accurate co-optimization of shape and control for many applications including vehicle design and soft robot optimization.

\section*{Acknowledgements}
We are grateful for funding received by the ETH AI Center and the Defense Advanced Research Projects Agency.


\bibliography{zotero_references}
\bibliographystyle{icml2022}

\newpage
\appendix
\onecolumn
\section{Detailed Computation Graph}
\begin{figure}[ht]
    \begin{center}
    \centerline{\includegraphics[width=1.3\columnwidth]{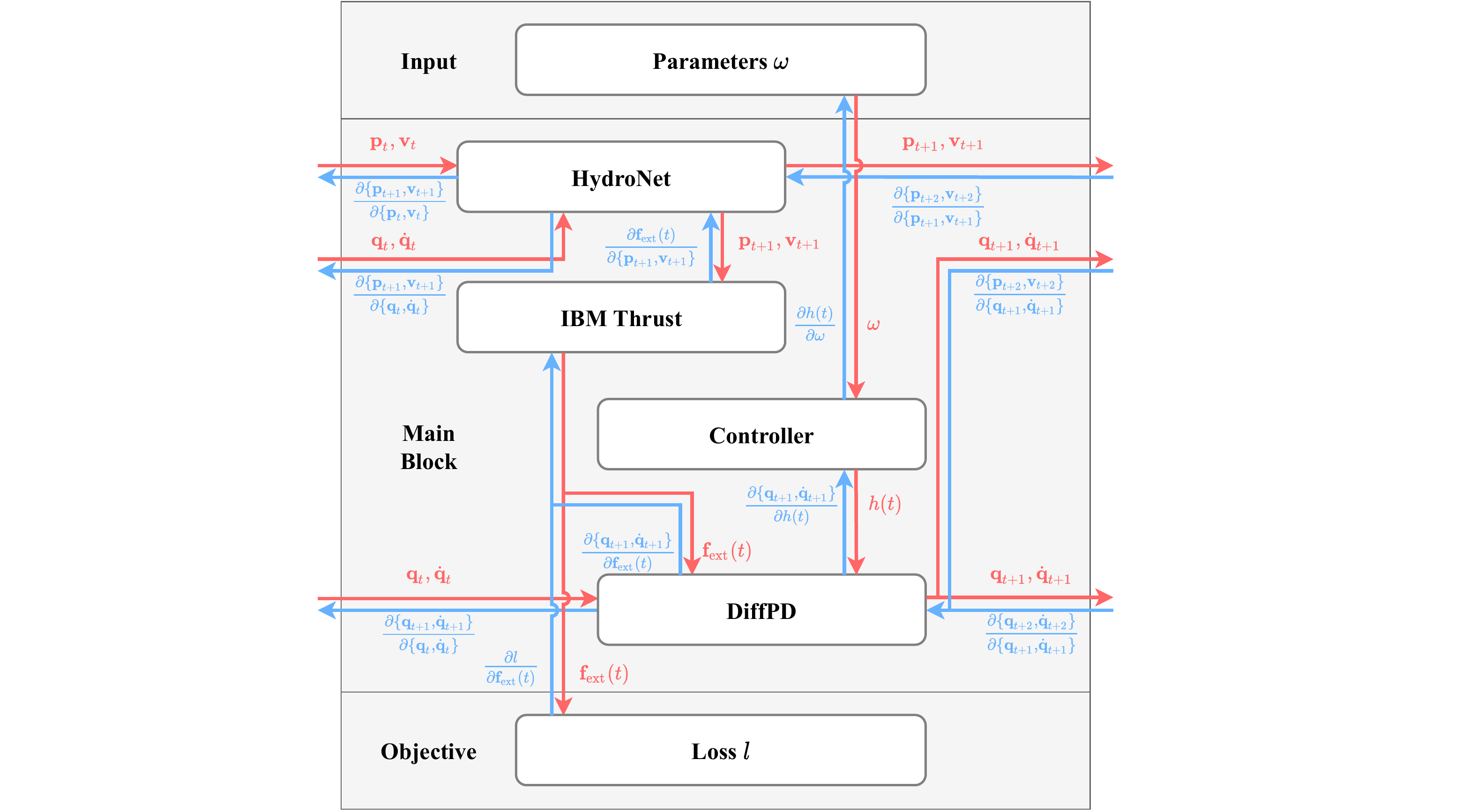}}
    \caption{Detailed computation graph for our simulator. Red arrows represent forward pass computation, while blue arrows represent backward pass computation of gradients of the block's variable with respect to forward pass quantitites.}
    \label{fig:sim-graph}
    \end{center}
    \vskip -0.2in
\end{figure}

\section{Fish Parametric Actuation}\label{appendix:fish}

\begin{figure}[ht]
\vskip 0.2in
\begin{center}
\centerline{\includegraphics[trim=0 0 0 0, clip,width=0.5\columnwidth]{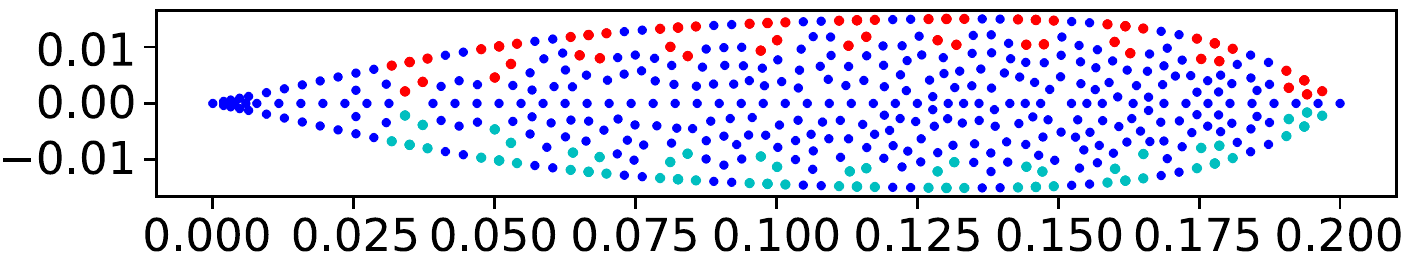}}
\caption{Discretized elements and actuated elements in the top half (red), and the bottom half (light blue) for the soft fish.}
\label{fig:mesh-carangiform}
\end{center}
\vskip -0.2in
\end{figure}

As shown in Figure~\ref{fig:mesh-carangiform}, we specify spaced elements on the fish surface to be actuated by contraction, with each actuated element in longitudinal position $X$ receiving an actuation signal $h(X,t)$ at time $t$. The carangiform style of swimming can be described by the following $h(X,t)$ with a backward-traveling wave and envelope:
\begin{equation}
    h(X,t) = C v(X) \sin(\gamma X + \omega t) (1-e^{-\frac{t}{t_a}}),
\end{equation}
where $C$ is a constant, $v(X)$ is the envelope of maximum lateral displacement, $\gamma$ the wave number of body undulations, $\omega$ the angular frequency, $c_0=\omega/\gamma$ the wave speed, and $t_a$ is the activation time. The envelope function $v(X)$ is given by
\begin{equation}
    v(X) = \frac{4}{25L}X^2 - \frac{6}{25}X + \frac{1}{10}L,
\end{equation}
where $L$ is the length of the fish, which for our case is $L = \SI{0.2}{\metre}$.

\section{The COMSOL Material Model}\label{appendix:comsol}
The fish is modeled as an isotropic and linear elastic solid in COMSOL Multiphysics. The Piola-Kirchoff stress (or engineering stress) is
\begin{equation}
    \mathbf{S}^e = 2 \mu_s \mathbf{E}^e + \lambda_s \textrm{tr} (\mathbf{E}^e) \mathbf{I},
\end{equation}
where $\mu_s$, $\lambda_s$ are the Lamé moduli and $\mathbf{E}^e$ is the Green-Lagrange strain. We can use an additive decomposition of the strain as
\begin{equation}
    \mathbf{E} = \mathbf{E}^e + \mathbf{E}^m, \label{eq:E_decomp}
\end{equation}
where $\mathbf{E}^m=\mathbf{E}^m(X,Y,t)$ is a time-evolving distortion strain that models the actuation of muscles. The Green-Lagrange strain can be written in terms of the deformation gradient as
\begin{equation}
    \mathbf{E} = \frac{1}{2}(\mathbf{F}^\top \mathbf{F} - \mathbf{I}).
\end{equation}
The deformation gradient permits the multiplicative decomposition
\begin{equation}
    \mathbf{F} = \mathbf{F}^e\mathbf{F}^m.
\end{equation}
\textbf{Note:} by defining the elastic deformation as (\ref{eq:E_decomp}), the elastic deformation is
\begin{equation}
    \mathbf{E}^m = \frac{1}{2}{\mathbf{F}^{m}}^\top ({\mathbf{F}^e}^\top\mathbf{F}^e - \mathbf{I})\mathbf{F}^m.
\end{equation}

To generate a flexural motion, we relate the distortion $\mathbf{E}^m$ to the curvature and the lateral displacement
\begin{equation}
    E^m_{xx}(X,Y,t) = -Y\frac{\partial^2 h(X,t)}{\partial X^2},
\end{equation}
where $Y=h(X,t)$, with $h(X,t)$ being the actuation scheme described in Section \ref{experimental-carangiform}, and the other components of $\mathbf{E}^m$ are identically zero.

\section{COMSOL Comparison plots}

The following plots qualitatively illustrate our comparison between our approach and the COMSOL simulation.

\begin{figure*}[ht]
\centering
\begin{minipage}[b]{0.49\textwidth}
    \raisebox{-0.75\height}{\includegraphics[width=\textwidth]{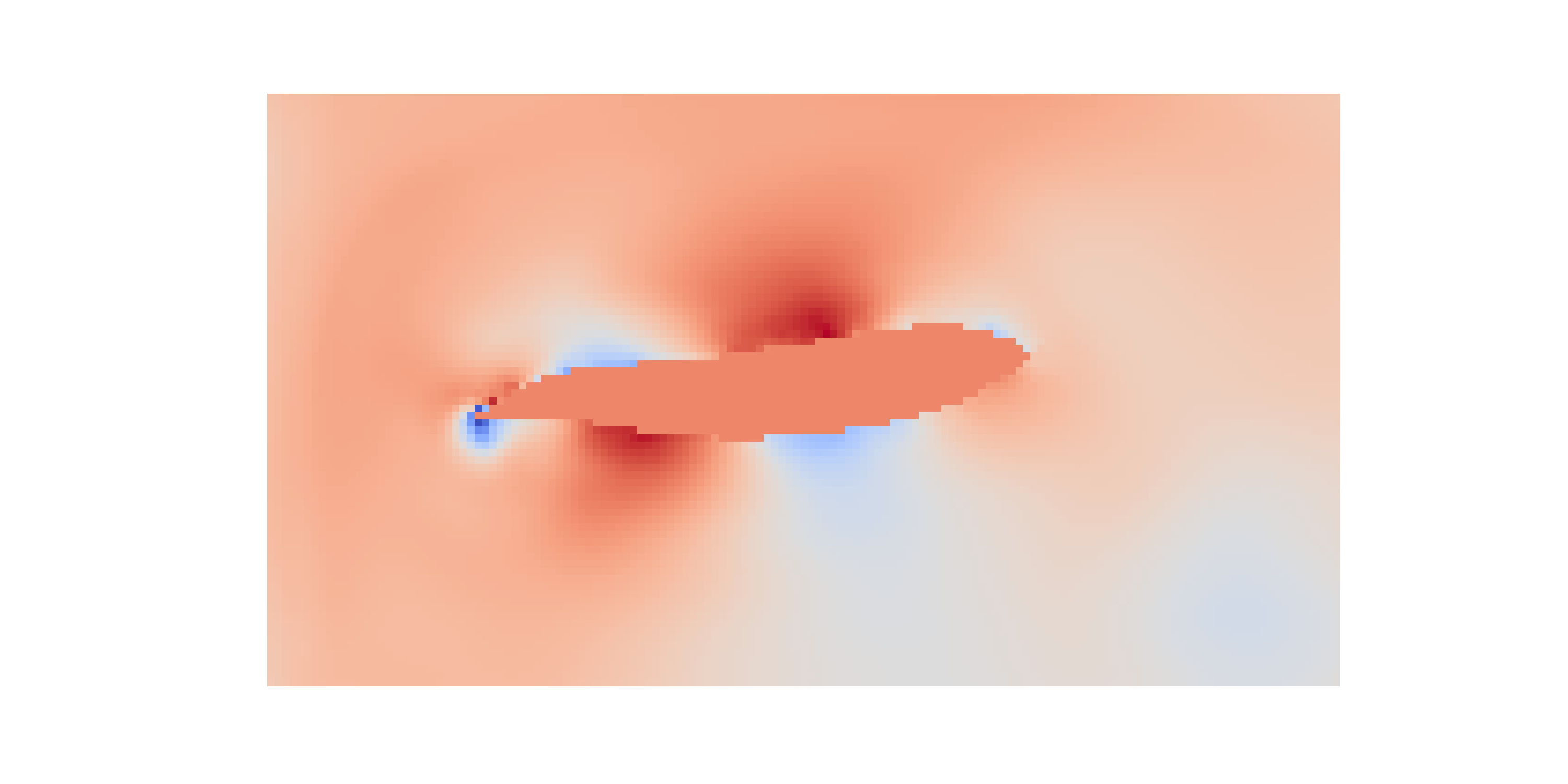}}
    \caption{Ours: DiffPD + HydroNet swimmer simulation.}
\end{minipage}
\begin{minipage}[b]{0.49\textwidth}
    \raisebox{-0.75\height}{\includegraphics[width=\textwidth]{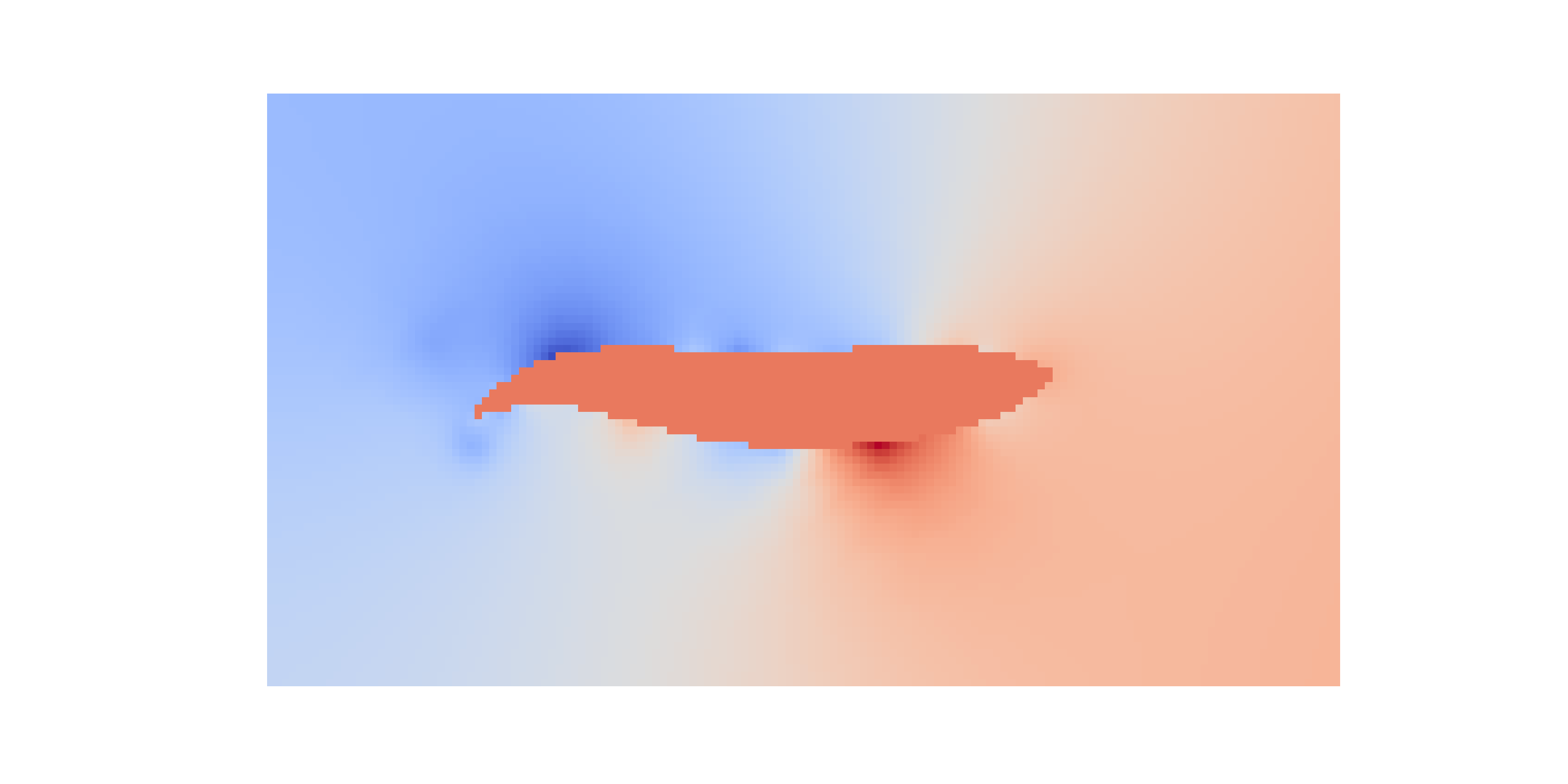}}
    \caption{COMSOL swimmer simulation.}
\end{minipage}
\caption{Normalized pressure fields for a frame of the two simulations.}
\label{fig:diffpd_sim}
\end{figure*}

\begin{figure*}[ht]
\centering
\includegraphics[width=\textwidth]{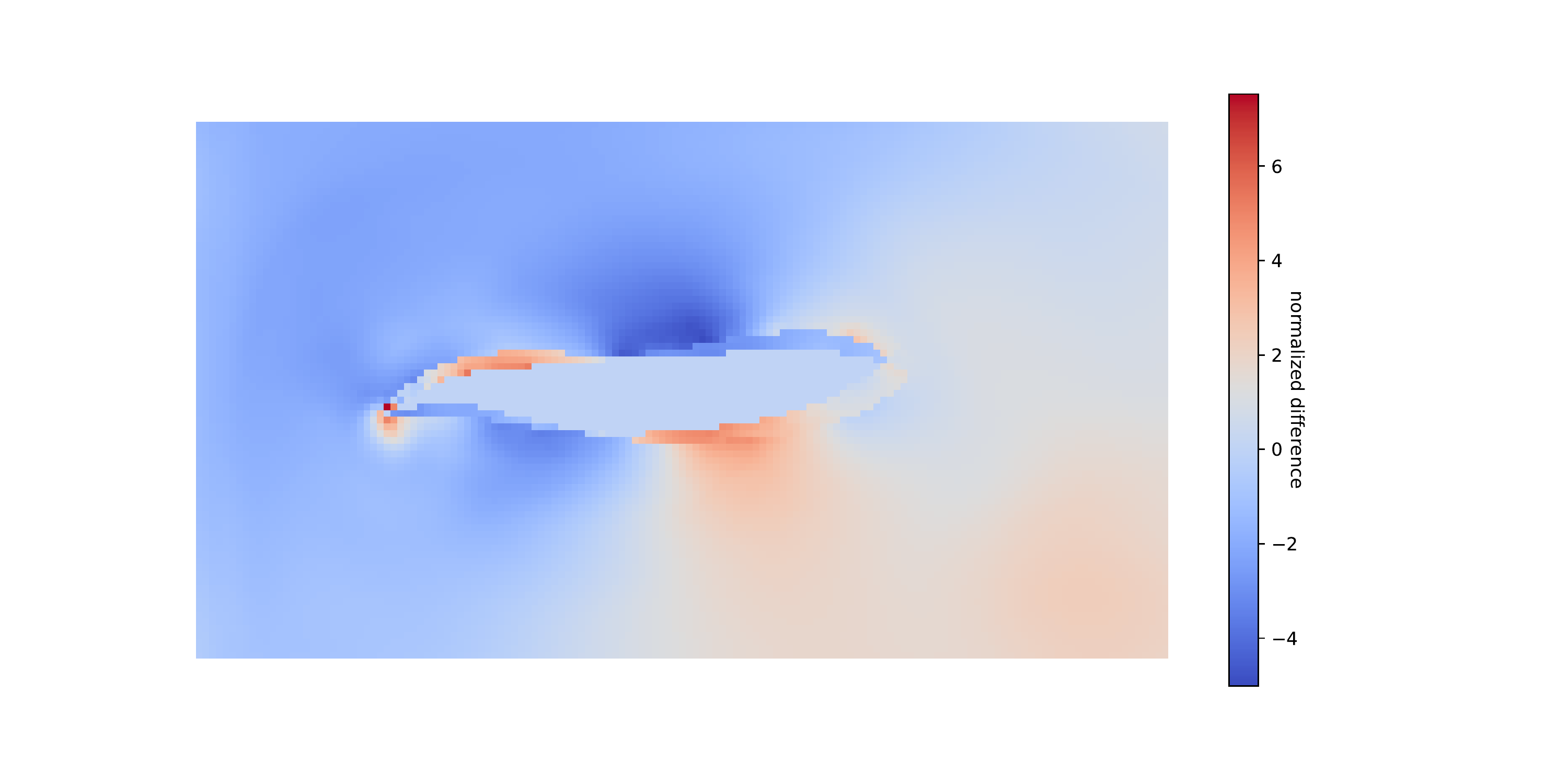}
\caption{Overlay of the normalized pressure fields of our DiffPD + HydroNet simulation and the COMSOL simulation.}
\label{fig:diff_sim}
\end{figure*}

\end{document}